\documentclass{article}
\usepackage{hyperref}
\pdfoutput=1
\usepackage{spconf,amsmath,graphicx}
\usepackage{multirow}
\usepackage{amsmath}
\usepackage{amsmath,amssymb}

\let\OLDthebibliography\thebibliography
\renewcommand\thebibliography[1]{
  \OLDthebibliography{#1}
  \setlength{\parskip}{0pt}
  \setlength{\itemsep}{1pt plus 1ex}
}
 
\title{Cogni-Net: Cognitive Feature Learning through Deep Visual Perception}
\name{Pranay Mukherjee\textsuperscript{1}, Abhirup Das\textsuperscript{1}, Ayan Kumar Bhunia\textsuperscript{2}\sthanks{Corresponding Author}, Partha Pratim Roy\textsuperscript{3}}
\address{\textsuperscript{1}Institute of Engineering \& Management, India \hspace{0.1cm} \textsuperscript{2}Nanyang Technological University, Singapore\\ \textsuperscript{3}Indian Institute of Technology Roorkee, India \\
{\tt\small \textsuperscript{2}ayanbhunia@ntu.edu.sg }
}
\begin{document}
%\ninept
%
\maketitle
\begin{abstract}
Can we ask computers to recognize what we see from brain signals alone? Our paper seeks to utilize the knowledge learnt in the visual domain by popular pre-trained vision models and use it to teach a recurrent model being trained on brain signals to learn a discriminative manifold of the human brain's cognition of different visual object categories in response to perceived visual cues. For this we make use of brain EEG signals triggered from visual stimuli like images and leverage the natural synchronization between images and their corresponding brain signals to learn a novel representation of the cognitive feature space. The concept of knowledge distillation has been used here for training the deep cognition model, CogniNet\footnote{The source code of the proposed system is publicly available at {https://www.github.com/53X/CogniNET}}, by employing a student-teacher learning technique in order to bridge the process of inter-modal knowledge transfer. The proposed novel architecture obtains state-of-the-art results, significantly surpassing other existing models. The experiments performed by us also suggest that if visual stimuli information like brain EEG signals can be gathered on a large scale, then that would help to obtain a better understanding of the largely unexplored domain of human brain cognition.
\end{abstract}

\begin{keywords}
Knowledge-distillation, Teacher-Student network, EEG Signal, Knowledge Transfer.
\end{keywords}
\section{Introduction}\label{sec:intro}

Recent advancements in computer vision have automated image classification tasks owing to complex deep convolutional neural networks and the availability of large scale labeled datasets like ImageNet. However, all of these CNN architectures attempt to learn a discriminative feature space that is data-dependent and limited to variances observed within the training dataset; they are not able to generalize well under unseen, irregular circumstances.
% The initial layers in such stacked convolutional architectures learn features relevant to , information in the deeper layers contain dataset-specific information.
For humans, the process of visual recognition amalgamates our sense of visual perception of attributes like shape and texture as well as cognition, the latter being an untapped source of information in automated image classification. Our work attempts to learn the cognitive-activity feature-space by utilizing these very signals (EEG) generated by humans during cognition and use them to discriminate between visual object categories. Works such as \cite{das2010predicting, wang2012combining, shenoy2008human, carlson2011high, carlson2013representational, kaneshiro2015representational, simanova2010identifying} show that brain signals evoked by a specific stimuli contain information about image categories, with \cite{spampinato2017deep} proving that they can be exploited to discriminate between image classes. Unlike \cite{spampinato2017deep}, our work attempts to utilize the link between the visual stimuli (images) and cognitive activity (EEG signals), taking inspiration from \cite{aytar2016soundnet} which attempts to learn acoustic representations from unlabeled videos. The authors of \cite{spampinato2017deep} used EEG signals for image classification, for which we already have very accurate deep learning models. Furthermore storing images for classification is very costly. In contrast to that, CogniNet allows the use of just milliseconds of recorded brain activity for classification of the already seen visual data thus performing the same task with reduced data while also being faster. Since we transfer knowledge from pre-trained close-to-perfect deep vision models to our network, it is possible to boost its performance horizon by using larger signal datasets. 
% Our deep signal classification network leverages the natural synchronization between images and their corresponding brain signals by being guided through a state-of-the-art vision model and hence achieves state-of-the-art performance for brain-signal classification. 
\begin{figure*}
    \centering
    \includegraphics[width=0.8\textwidth]{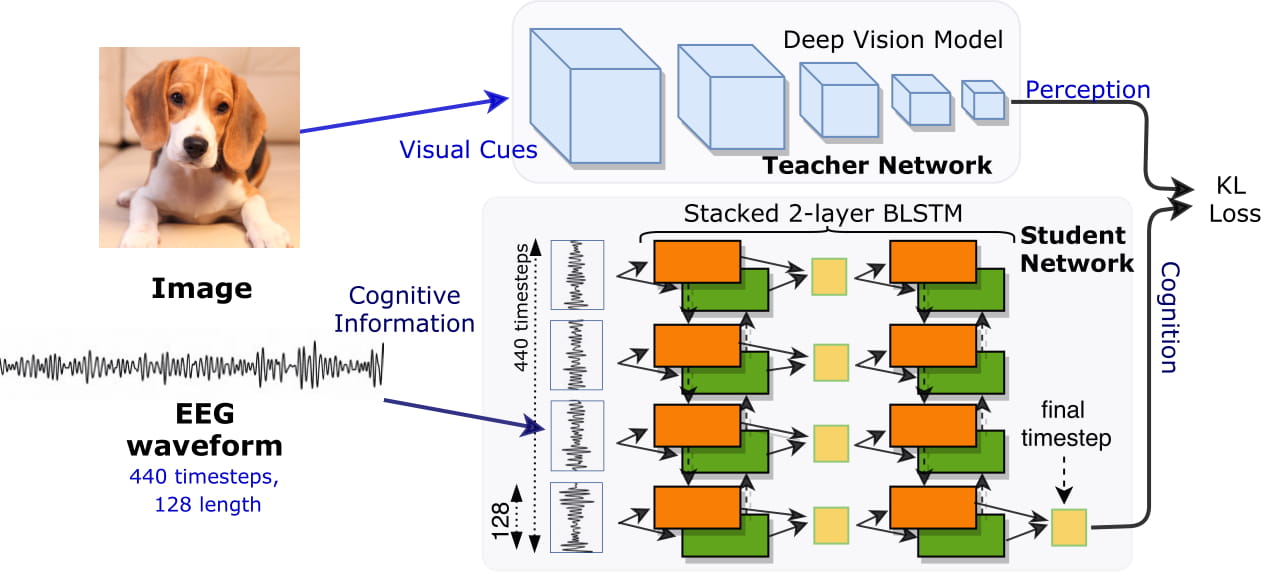}
    \caption{\textbf{CogniNet}: We propose a  deep recurrent architecture comprising BLSTM units for brain EEG signal recognition. The proposed approach makes use of knowledge distillation to transfer visual discriminative knowledge from the teacher network $\mathcal{T}$, a pre-trained image classification model, to the student network $\mathcal{S}$ which is a deep 2-layer BLSTM model.}
    \label{fig:my_label}
\end{figure*}

While transfer learning has been extensively studied in computer vision and has been successfully used for tasks like image segmentation and object detection \cite{aytar2011tabula, aytar2015part}, our work makes use of cross-modal knowledge transfer which has also been addressed by authors of \cite{ngiam2011multimodal, huang2013audio, castrejon2016learning, owens2016visually}. Consequently, another driving idea on which our work is based on is that of teacher-student networks \cite{ba2014deep, gupta2016cross} and the transfer of dark knowledge [21] between two different models. The dark knowledge transfer successfully compresses the required discriminative knowledge from a cumbersome, complex and deep model to a comparatively lighter model without the considerable degradation in performance but with the added benefit of being able to deploy it more efficiently under real-time circumstances. Though previous works [16, 21] have used this technique for the transfer of knowledge within the same modality, the authors of \cite{aytar2016soundnet} has applied the same technique for cross-modal knowledge transfer with considerable success. Here, we have also applied the concept of cross-modal transfer of knowledge between the teacher and student networks where the teacher network operates on vision to train the student model which operates on EEG brain signals.

Our work capitalizes on the natural synchronization between images and EEG signals that are evoked by visual object stimuli to learn a deep discriminative representation of these brain signals under both supervised and unsupervised settings. For the task at hand, the teacher network  is a  state-of-the-art image classification model pre-trained on ImageNet with the student network being a deep Recurrent Neural Network model (stacked BLSTM). Thus in our case, the deep convolutional vision network teaches the deep recurrent network to correctly classify brain signals. Although we have used a vision network as the teacher network, our student model doesn't depend on visual cues during the time of inference due to the fact that the knowledge distillation process helps us to transfer the dark knowledge present in the wide and deep vision model to a narrower and shallower recurrent model. 

Our contribution through this work is three fold. First, to the best of our knowledge, we make a novel contribution using knowledge distillation for brain signal classification. Second, the proposed brain signal network, CogniNet, when trained under unsupervised settings acts as a novel feature extractor and shows competitive performance while handling unseen signal categories. Thirdly, we also show that when trained under supervised settings, our model easily beats the current state-of-the-art algorithm for brain signal classification.

\section{Proposed Methodology}
\label{sec:approach}
The principle idea of this work is to utilize the knowledge of well trained deep image classification models to help train our recurrent model, CogniNet, to learn discriminative feature representations of brain-signal data. We elaborate on the two networks involved in this process of knowledge transfer. 

\textbf{Teacher Vision Network}
The knowledge distillation technique [21] makes use of a semantically rich pre-trained image classification model. We experiment with some popular state-of-the-art deep learning models (Section 3) formed of strided convolutional layers that accept images to give corresponding feature representations. Such a vision-based model like \cite{krizhevsky2012imagenet,szegedy2015going,simonyan2014very} acts as the teacher network $\mathcal{T}$. 

\textbf{Student Recurrent Model}
We use a deep recurrent model $\mathcal{S}$ comprising of BLSTMs as our student network for learning the brain EEG signal representations and classifying them. Deep recurrent models are suitable for processing brain EEG signals mainly because they are time-varying signals. Furthermore, like images, we want our signal network to be translation invariant. Significantly, stacked BLSTM layers which work surprisingly well under complex conditions \cite{spampinato2017deep} help in obtaining a high-level encoding of features from within these raw brain signals. To tackle the problem of varying temporal lengths of the brain signals, we follow the work proposed in \cite{spampinato2017deep} and represent every input brain EEG signal as a matrix, $M \in \mathbb{R}^{440\times128}$ where the rows represent the effective $440$ timesteps and the columns represent the $128$ channels \cite{spampinato2017deep} at each timestep.

\textbf{Loss Function }
For every pair $\{x_i, y_i\}$ of EEG signals and images, we make usage of the posterior probabilities $\mathcal{T}(y_{i})$ given as output by the teacher network to train the student network so that the latter is able to recognize the visual concepts present in brain EEG signal $x_i$ and output the predicted probabilities $\mathcal{S}(x_{i})$. Thus the teacher $\mathcal{T}$ transfers image based concepts to the student $\mathcal{S}$. Considering the outputs of the teacher model to be a distribution of classification targets, this task of knowledge transfer uses KL-divergence loss, denoted by $D_{KL}$, to construct the loss function that we aim to minimize during training. Thus, the objective function is given by the following equation:
\begin{align}
\label{divergence_loss}
L_{div}\ =\ \sum_{i=1}^N \ D_{KL}\ (\mathcal{T}(y_{i})\ ||\ \mathcal{S} (x_{i};\mathbf{\theta}))    \\
\mathrm{where}, \ \label{KL_divergence}
D_{KL}( P\ ||\ Q)\ =\ \sum_j^N P_j\log \frac{P_j}{Q_j} 
\end{align}
We optimize the loss in eqn. \ref{divergence_loss} through back propagation using stochastic gradient descent since KL-divergence is differentiable. 
% Once the brain EEG network has been trained we measure it's performance using the accuracy metric on the testing set. 
It is  worth noting that the above mentioned training of the brain EEG student network using knowledge distillation method is done under both unsupervised setting and supervised settings for separate experiments.  Under supervised settings, we minimize a weighted summation of $L_{div}$ and the categorical cross entropy loss function $L_{cat}$ as per the original paper \cite{hinton2015distilling}. Thus under supervised settings we minimize the following:

\begin{align}
\label{eq:total_loss}
&\mathrm{L_{opt}} =  \frac{\mathrm{T}}{2}^2 L_{div} + 0.5L_{cat} \\
\mathrm{where}&, \ L_{cat}(P,Q) = \frac{1}{N} \sum_j^N P_j \log Q_j
\end{align}

Consequently, under unsupervised settings, the objective term that we minimize consists of only the KL-divergence loss, $L_{div}$. During inference, the student network has zero dependence on vision-based cues. Thus in this work, we project images into a completely new cognitive manifold so that EEG-based discriminative features can be better learned. 

\textbf{Classifying unseen categories}
Under real world conditions it is quite natural to encounter out-of-vocabulary (OOV) signals that the CogniNet has not been trained on. To tackle these categories, we use a unsupervisedly trained (using the KL divergence loss only) CogniNet as a feature extractor for fitting standard machine learning classifiers like SVMs on a small dataset of labelled brain EEG signals containing the classes of interest. A data augmentation method was followed for ensuring the robustness of our model where every brain EEG signal was divided into overlapping excerpts of a suitable length and the feature extraction was performed on each of these excerpts using the CogniNet. The final predictions on the original EEG signal was finally made using majority voting. 

\section{Experiments}
\label{sec:exp}

The implementation for our approach is done using the Keras deep learning library. For the experiments, we have minimized the loss functions using the Adam optimizer and by fixing the learning rate and momentum at 0.001 and 0.9 respectively. The batch size of 32 was found to work the best for the conducted experiments.

\textbf{The Dataset}
For our experiments we use the dataset released in \cite{spampinato2017deep}. This visual stimuli dataset is a small subset of the ImageNet dataset with 40 classes with 50 images per class and accompanying EEG signal data. Temporal information was collected from each signal sample following the work proposed in \cite{spampinato2017deep}. Each of 6 different subjects were made to see the 2000 images to generate 12000 EEG-signal samples. The recorded brain activity data includes Gamma and Beta bands \cite{spampinato2017deep} that convey information about active cognitive processes. The obtained dataset is split according to the train-test split ratio of 70:30 in such a manner that all the 6 waveforms (obtained from 6 different subjects) for any particular image belongs to either the training set or the test set but not both.

\textbf{Performance Analysis}
In this section we evaluate the performances of a few different baselines with different training methodologies and compare them with the work proposed here. As the first baseline, the pairs of EEG signals and corresponding target values are taken and fed into the a 2-layer stacked BLSTM network followed by a fully-connected layer with softmax activation. This model is trained in a supervised manner without the help of any teacher network. For the second baseline, we used the ``common LSTM + output layer" model described in \cite{spampinato2017deep}.
We use the 1D convolutional encoder model proposed in \cite{aytar2016soundnet} as the third baseline because of it's history of successfully capturing temporal acoustic information. All of the above baselines we trained by minimizing $L_{cat}$ function. Since our dataset is balanced we use accuracy as the performance measure for all the experiments. Standard machine learning classification algorithms like SVM, Decision trees, Random Forest etc. are incompetent in processing the raw brain EEG signals and hence are not used as baselines. 

\begin{table}
\parbox{\linewidth}{
\centering
\begin{tabular}{cc}
\hline
\textbf{Method} & \textbf{Accuracy}\\
\hline
2-layer BLSTM & 0.832\\
common LSTM + output layer \cite{spampinato2017deep} & 0.811\\
1D-Convoltional network \cite{aytar2016soundnet} & 0.804\\
2-layer BLSTM + VGG-19 & 0.875\\
2-layer BLSTM + AlexNet & 0.883\\
\textbf{2-layer BLSTM + GoogleNet}  & \textbf{0.896}\\
\hline
\end{tabular}
\caption{We evaluate the classification accuracy on the test-set for different models. For the last three models in the table, we minimized $\mathrm{L_{opt}}$ (eqn. \ref{eq:total_loss}).}
\label{tab: exp1}
}
\end{table}
 
Performance on unseen image categories are shown on a subset of the original brain signal dataset consisting of signals belonging to $6$ image classes. For this, the CogniNet was first trained in an unsupervised manner using the images and signals belonging to the remaining $34$ categories. The features are then extracted from it's final recurrent layer on top of which an SVM (or any other classifier) is trained in a supervised manner. We tested with 64, 128, 256 and 512 dimensional feature vectors. For all the cases, 128 dimensional vectors performed the best followed by 256, 64 and 512 dimensional ones. We reason that this occurs due to overfitting.
\begin{table}
\centering
\begin{tabular}{ccccc}
\hline
\textbf{Method} & \textbf{SVM} & \textbf{kNN} & \textbf{RF}\\
\hline
2-layer + VGG-19 & 0.745 & 0.683 & 0.734\\
2-layer + AlexNet & 0.767 & 0.701 & 0.751\\
\textbf{2-layer + GoogleNet} & \textbf{0.781} & 0.725 & 0.772\\
\hline
\end{tabular}
\caption{Performance comparision of various ML classifiers fitted on top of the 128 dim. feature vectors extracted from the final BLSTM layer of the CogniNet trained using GoogleNet as the teacher network under unsupervised settings.}
\label{tab: exp1}
\end{table}
\begin{table}
\centering
\begin{tabular}{ccc}
\hline
\textbf{Comparison of} & \textbf{CogniNet} & \textbf{Accuracy}\\
\hline
\multirow{2}{*}{Loss} & 2 Layer, $\ell2$ & 0.725\\
& \textbf{2 Layer, KL} & \textbf{0.896}\\ 
\hline
\multirow{4}{*}{Depth} & 1 Layer, KL & 0.836\\
& \textbf{2 Layer, KL} & \textbf{0.896}\\
& 3 Layer, KL & 0.802\\
& 4 Layer, KL & 0.738\\
\hline
\end{tabular}
\caption{A comparative study for the accuracy of the various configurations of CogniNet. For all the cases, we trained the  CogniNet using GoogleNet under supervised settings.}
\label{tab: exp1}
\end{table}

\textbf{Ablation Analysis}
 We perform ablation analysis for the better comprehensibility of our proposed work.
 
The first ablation study that we performed was on the loss function used for the training the student network under the supervision of the teacher network for supervised settings. For this we used $\ell 2$ regression loss instead of $L_{div}$ in eqn. \ref{eq:total_loss} and got worse results.  
%INSERT the Table 3 here.(3 X 3)
\begin{table}
\centering
\begin{tabular}{ccccc}
\hline
\textbf{Method} & \textbf{$\mathrm{T}=1$} & \textbf{$2$} & \textbf{$5$} & \textbf{$10$}\\
\hline
2-layer + VGG-19 & 0.866 & 0.872 & 0.875 & 0.869\\
2-layer + AlexNet & 0.871 & 0.877 & 0.883 & 0.873\\
\textbf{2-layer + GoogleNet} & 0.884 & 0.892 & \textbf{0.896} & 0.887\\
\hline
\end{tabular}
\caption{Accuracy comparision of our models as the value of T (temperature parameter) changes}
\label{tab: exp1}
\end{table}

Secondly, we studied the effect of the student network's depth on the performance. We compare the performance of the 2-layer CogniNet with 1-layer, 3-layer and 4-layer CogniNet architecture. It was found that the performance of the 2-layer CogniNet with a recurrent dropout of 0.5 is better than that of the single layer CogniNet proving that the deeper networks helps in the better understanding of brain signals. However the performance dips for 3-layer CogniNet and 4-layer CogniNet the reason for which being overfitting. However, the 1-layer CogniNet still performs better than the existing state-of-the-art methods on the domain.

In the third ablation study, we examine the importance of the unlabeled image data to training of the CogniNet, for which we look at the first baseline , where the model is trained with respect to available labelled signal data in a supervised manner, without utilizing any images. This baseline thus allows us to quantify the effective contribution of these unlabeled images. The 2-layer BLSTM without unlabelled data has higher accuracy than 1-layer, but performs significantly worse than 2-layer CogniNet that is trained with unlabeled image data using $L_{opt}$ loss thus signifying the importance of $L_{div}$ (KL-divergence loss) in our model. This means that the unlabeled visual cues significantly complement our model's ability to learn the EEG signal feature space.

Finally, we provide the variation in performance of our model as the value of the temperature parameter, $T$ is varied in  $L_{opt}$ (eqn. \ref{eq:total_loss}). For this we choose the values of $T$ as 1, 2, 5, 10 as per the original work. The comparative results are shown in Table 4.
\begin{figure}
    \centering
    \includegraphics[width=0.8\linewidth]{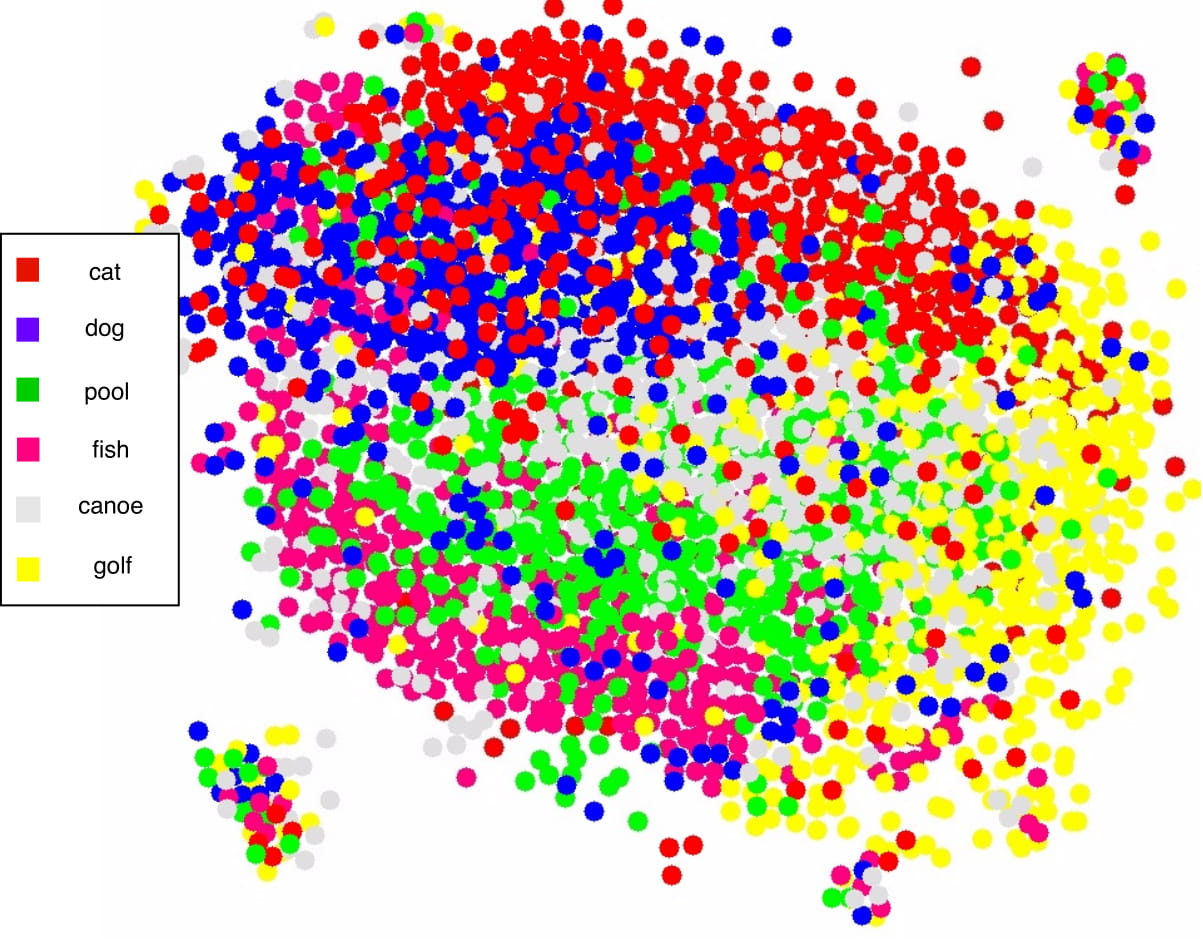}
    \caption{A t-SNE plot of EEG signals for 6 selected ImageNet classes with features gathered from the second BLSTM layer of the proposed CogniNet model.}
    \label{fig:my_label}
\end{figure}

\section{Conclusion}

In this work, we train the CogniNet by using the natural synchronization between images and their corresponding brain EEG signals that are triggered when images act as stimuli to the visual cortex. Our work uses the knowledge-distillation technique by selecting a vision network as the teacher and the CogniNet as the student to model the semantically rich features of the obtained brain signals. Our results provide strong evidence that images can be leveraged for learning powerful and robust brain signal representations with the aid of state-of-the-art teacher networks. Also by increasing the amount of gathered data, we can obtain an even richer understanding of the domain of brain signals by using machine vision.

\bibliographystyle{IEEEbib}
\bibliography{strings,refs}

\end{document}